**Title:** Shape-programmable Adaptive Multi-material Microrobots for Biomedical Applications

**Authors:**
Liyuan Tan,[1]* Yang Yang,[1] Li Fang,[2] David J. Cappelleri[1,2]

**Affiliations:**
[1]School of Mechanical Engineering, Purdue University, West Lafayette, IN 47907, USA.
[2]Weldon School of Biomedical Engineering, Purdue University, West Lafayette, IN 47907, USA.

*Corresponding author. Email: tan328@purdue.edu.

**Abstract:** Flagellated microorganisms can swim at low Reynolds numbers and adapt to changes in their environment. Specifically, the flagella can switch their shapes or modes through gene expression. In the past decade, efforts have been made to fabricate and investigate rigid types of microrobots without any adaptation to the environments. More recently, obtaining adaptive microrobots mimicking real microorganisms is getting more attention. However, even though some adaptive microrobots achieved by hydrogels have emerged, the swimming behaviors of the microrobots before and after the environment-induced deformations are not predicted in a systematic standardized way. In this work, experiments, finite element analysis, and dynamic modeling are presented together to realize a complete understanding of these adaptive microrobots. The above three parts are cross-verified proving the success of using such methods, facilitating the bio-applications with shape-programmable and even swimming performance-programmable microrobots. Moreover, an application of targeted object delivery using the proposed microrobot has been successfully demonstrated. Finally, cytotoxicity tests are performed to prove the potential for using the proposed microrobot for biomedical applications.

**One-Sentence Summary:** A systematic approach to design shape-programable, dual-function, and adaptive microrobots for biomedical applications.

**Main Text:**

**INTRODUCTION**

Microorganisms are capable of swimming with flagella to provide motility (*1–3*). These microorganisms can adapt their flagella into different shapes or modes by altering gene expression to accommodate environmental changes or for other proposes like nutrition, hosting, and invasion (*4*). For example, the flagella of a spermatozoon of *Echinus esculentus* will result in a transition from a planar to a helical shape when the viscosity is increased and back to a quasi-planar shape when it is further increased (*5*). Moreover, recent investigations show that the flagella can deform to wrap around the cell body to escape from traps or to enhance the efficiency of environmental exploration (*6, 7*).

Inspired by these natural living beings, many microrobots have been fabricated to swim in this microscale world. The two strategies most adopted to achieve motility are the helical structures mimicking the flagella of bacterial *E. coli* and the flexible body replicating the motion of a spermatozoa (*8*). In the last decade, various helical-type microrobots are realized with fixed shapes, i.e., the structure will not change once it is fabricated (*9–11*). These designs typically also have magnetic properties incorporated in them so they can be controlled via rotating magnetic fields. However, the swimming performance of these microrobots is predictable sharing an identical normalized frequency response for certain geometry and magnetic moment (*12*). In this case, the dominant parameter is the viscosity which limits and lowers the highest achievable velocity when the viscosity is increased. Therefore, it is critical to manufacture microrobots that are allowed to modify themselves to environments to evaluate and learn from the microorganisms.

The recent development of utilizing smart materials enables us to develop microrobots with adaptive functionalities in order to achieve multiple behaviors like a flagellated microorganism. Microrobots obtained via these smart materials are able to deform their shapes upon the stimulation of environmental cues, such as temperature, pH, and osmolarity changes (*13*). In those smart materials, hydrogels are one of the most adopted materials for microrobots due to their outstanding biocompatibility. Based on these hydrogel materials, many adaptive microrobots are produced with the capability of adapting shapes to environmental signals, passing through confined channels, and targeted drug-delivery (*14–17*).

Even though there are many investigations on deformations of structures made of hydrogels using finite element analysis (FEA), including helical deformations, those simulations are mostly qualitative or quantitative with some classical material parameters (*18–22*). Additionally, they investigate structures at a larger scale as the material properties of the hydrogel at the microscale are not easily accessible, since characterization tests would require the hydrogel to stay in a corresponding solvent (*23*, *24*). Therefore, the prediction of swimming performances of an adaptive microrobot before and after stimulation is still under debate while dynamic modeling requires known geometrical parameters and material properties.

In this work, we propose a systematic standardized method for predicting the transition of swimming performances of an adaptive multi-material microrobot upon environmental stimulation. The method is realized through material property tests, FEA, and dynamic modeling. While the environment-stimulated deformation is predicted by FEA, the difference in swimming performance is calculated by dynamic modeling. Both the deformation simulation and dynamic calculation are verified experimentally with 3D-printed tail structures and multi-material microrobots made by combining traditional microfabrication techniques and 3D printing through two-photon polymerization. A functional end-effector is also printed with the microrobot for micromanipulation tasks. Specifically, targeted object delivery of spheres is demonstrated with our multi-material microrobot. This work provides a versatile method to study shape-programmable adaptive multi-material microrobots for bio-applications with different requirements of swimming performance with both advanced functionalities of adaptive locomotion and micromanipulation.

# RESULTS
## Simulation-assisted design and fabrication of the adaptive multi-material microrobots
The design overview of the proposed microrobots is presented in Fig. 1A. In general, the microrobot has three major components, including a magnetic main body patterned with SU-8 photoresist and magnetic microparticles (MMPs), a helical tail achieved by shape adaptive hydrogels with soft and hard modulating regions, and a functional end-effector printed with IP-S for manipulation tasks. Except for the SU-8 and SU-8/MMPs composite that are fabricated via photolithography, the end-effector (a holder) and the helical tail are both 3D-printed by TPP. The schematic of printing the adaptive hydrogels is shown in Fig. 1B. The monomer *N*-Isopropylacrylamide (NIPAM) is co-polymerized with acrylic acid (AAc) with dipentaerythritol pentaacrylate (DPEPA) as crosslinker. With the photoinitiator *4,4'*-bis(diethylamino)benzophenone (EMK), the precursor is crosslinked via the TPP process triggered by a femtosecond laser. The deformation degree of the printed hydrogels can be controlled by the dose of energy applied by adjusting the laser power (LP) and scanning speed (SS). Typically, a higher LP and slower SS will give a larger dose and therefore a higher crosslinking density with more polymer chains per volume and vice versa. Since the dose can be adjusted by either the LP or SS, the dose in this work is changed by the LP while the SS is held consistent. Fig. 1C shows the schematic of the printing of the helical tail. The helical structure is achieved by modulating between soft adaptive and rigid regions. By tuning the LP of different soft regions, various helical structures can be realized. For example, by increasing the LP gradually from one side to the other, a conical helical tail is obtained. The deswelling ratios of the hydrogels printed by different LPs are measured and presented in Fig. 1D. The ratio of deswelling ratio of water to IPA based on the experimental data is also provided for reference. The deswelling ratio in water is fitted with an exponential function with a limit of 1. However, the deswelling ratios at higher powers show a ratio higher than 1 which is most likely due to over-exposure at such powers.

To achieve shape programmable deformation of the microrobots, FEA simulations are carried out to obtain the geometries before and after the deformation in various environments. The key to realizing accurate simulation of the fabricated microrobots is the material properties of the corresponding hydrogel that are crosslinked with different LPs. Such properties can be obtained by performing uniaxial compression tests on test samples printed by different LPs. Typically, the material properties of a hydrogel at a certain equilibrium state can be characterized by three parameters: the crosslinking density ($Nv$), the initial stretch at the free-swelling state ($\lambda_0$), and the Flory interaction parameters ($\chi$) representing the interaction between the polymer network and the solvent. While the $\chi$ can be calculated from $Nv$ and $\lambda_0$ via Eq. S5, the $Nv$ and $\lambda_0$ can be obtained by fitting the experimental data of the compression tests according to (*24*) using the equation below:

$$Nv\left(\lambda_0\lambda_1' - \frac{1}{\lambda_0\lambda_1'}\right) + \lambda_0^2\lambda_2'^2\ln\left(1 - \frac{1}{\lambda_0^3\lambda_1'\lambda_2'^2}\right) + \frac{1}{\lambda_0\lambda_1'}$$
$$-\frac{Nv(\lambda_0^2-1)+\lambda_0^3\ln\left(1-\frac{1}{\lambda_0^3}\right)+1}{\lambda_0\lambda_1'^2\lambda_2'^2} = \frac{\lambda_0^2\sigma_1'}{M} \qquad (1)$$

with $M = kT/v$ and $\lambda_2'^2 = \lambda_1'^2 - \sigma_1'\lambda_0\lambda_1'/(\lambda_2'^2 MNv)$, where $Nv$ is the crosslinking density or chain number to the dry state, $\lambda_0$ is the stretch at the free-swelling state to the dry state, $k$ is the Boltzmann constant, $v$ is the volume of per solvent molecule, $T$ is the absolute temperature, and $\sigma_1'$ is the nominal stress. $\lambda_1$, $\lambda_2$, and $\lambda_3$ are the stretches to the dry state along three principal directions. For uniaxial compression tests, $\lambda_2$ equals $\lambda_2$ assuming that the samples are isotropic. $\lambda_1' = \lambda_1\lambda_0$ and $\lambda_2' = \lambda_2\lambda_0$ are stretches relative to the free-swelling state. $\sigma_1'$ and $\lambda_1'$ are obtained from the compression test. Therefore, only two parameters, $Nv$ and $\lambda_0$, are unknowns and they can be obtained by fitting the experimental data with Eq. 1.

The relation between the nominal stress $\sigma_1'$ and the stretch $\lambda_1'$ with respect to the free-swelling state is tested with uniaxial compression for every 1 μm by a custom-built testing system (see Supplementary Materials). Some representative test results are plotted in Fig.

1E. As can be seen in the plot, the absolute stress increased significantly from 26 mW to 30 mW while the difference from 30 mW to 60 mW is not as pronounced. Fig. 1F gives the results of $Nv$ and $\lambda_0$ obtained by fitting them with Eq. 1. Both the $Nv$ and $\lambda_0$ are leading to strong exponential trends. The $\chi$ values are calculated using Eq. S5 based on the individual and fitted $Nv$ and $\lambda_0$, respectively. As can be seen in Fig. 1G, the red data points are the $\chi$ obtained from individual $Nv$ and $\lambda_0$ and are showing non-trivial errors because of the limitations of the system. However, the blue line acquired from the fitted $Nv$ and $\lambda_0$ is consistent with the scatter values within a range higher than 0.75. Since the entire printed structure is deforming among three situations, initial shape as designed, deformed shape in IPA, and deformed shape in the water, the $\chi$ values in IPA and for maintaining the initial shapes are calculated according to the sizes in IPA and the size as designed, while having an identical $Nv$ for each crosslinking density. Based on the results in Fig. 1D to 1G, a saturation point is found at around 35 mW. As for $\chi$ values, the difference between the shape in water and the initial shape increases as the LP decreases. However, the $\chi$ values in IPA are overall smaller than in water by about 0.45. Test results in Fig. 1D to Fig. 1G are for soft regions with a slicing distance of 0.3 μm and a hatching distance of 0.2 μm with a saturation of $Nv$ of around 0.16 which is too weak to support a helical deformation. Therefore, smaller slicing and hatching distances of 0.2 μm and 0.1 μm are used, respectively, to obtain a stiffer material with a saturation of $Nv$ of around 0.25. Materials properties corresponding to the stiffer hydrogel are also measured and provided in Fig. S1.

Starting with the Helmholtz free energy function, the final form of the free energy function used for FEA implementation can be expressed as

$$W = \frac{1}{2} Nv \left[ \lambda_0^{-1} J'^{\frac{2}{3}} \bar{I}_1 - 3\lambda_0^{-3} - 2\lambda_0^{-1} \ln(\lambda_0^3 J') \right] \\ + (J' - \lambda_0^{-3}) \ln\left(\frac{J'}{J' - \lambda_0^{-3}}\right) - \frac{\chi}{\lambda_0^6 J'} - \frac{\mu}{kT}(J' - \lambda_0^{-3}) \tag{2}$$

after introducing a Legendre transform and using the free-swelling state as the reference state (21) with $J = \det \mathbf{F} = \lambda_1 \lambda_2 \lambda_3 = \lambda_0^3 J'$ and $\bar{I}_1$ being the invariants of the deformation gradient $\mathbf{F}$. Also, $J$ and $J'$ are the volume ratios to the dry and free-swelling states, respectively. The free energy function can be then implemented using the UHYPER subroutine with the commercial FEA software ABAQUS (Abaqus/CAE 2018, Dassault Systèmes S.A., France). Fig. 1H shows an example of a flower-like structure with six petals printed with different LP for the soft regions for different petals. For each petal, the LP of the corresponding soft regions is increased by 2.5 mW. Both experimental and simulation results are presented and show a decreasing bending degree of the petals with the increase of LP. The slight deviation between the experiment and simulation is mainly because of the measurement error of the material properties which show sharp changes in the power range between 20 mW and 35 mW and errors from fabrication. Nevertheless, they are in good agreement and are accurate enough for qualitative investigations.

**Programming the helical hydrogel tail**

The helical tails are achieved upon deformation from planar strips with a modulation of soft and hard regions. A demonstration of the planar tail strip and the design parameters are shown in Fig. 2A. While $W$ and $L$ are the width and length of the entire strip, the modulating angle $\theta$ represents the angle between the stripe orientation and the length direction of the entire strip. The helical deformation is driven by the deswelling of the soft regions which forms a bilayer structure with a supporting hard layer. The soft regions have a thickness of $h_1$, while the overall thickness of the strip is $h_2$. Therefore, the hard regions have a thickness of $h_2$ corresponding to the overall thickness, and the thickness of the supporting layer underneath the soft regions is $h_2 - h_1$. The parameters $b_1$ and $b_2$ correspond to the widths of the soft and hard regions. Fig. 2B illustrates the geometrical parameters of a deformed helical shape. In general, a helix can be defined by its diameter $D$, pitch $P$, helical angle $\alpha$, and number of turns $n$. In this paper, the soft regions are printed with an LP of 23 mW while the hard supporting layer is printed with an LP of 35 mW unless otherwise mentioned. The deformed helical tails with different modulating angles in IPA and water are shown in Fig. 2C and Fig. 2D, respectively, with simulation results. The simulations are calculated with the same geometrical parameters as used for printing the tails. Some fixtures are also printed for the tails so that the deformed tails can be fixed in place for further measurement with different solvents. The helical tails are more compact in water than in IPA since the soft regions deswell more in water as shown in Fig. 1D. The helical parameters, except for the helical angles, are measured from images with tails like those in Fig. 2C from multiple samples while the parameters from simulations are fit based on the resulting shapes shown in Fig. 2C as well. The helical angle is calculated with the relationship as below

$$\alpha = \arctan\left(\frac{\pi D}{P}\right). \tag{3}$$

The obtained geometrical parameters are collected and plotted in Fig. 2E to Fig. 2H in IPA and water. As can be seen from the figures, the pitch, diameter, and turns have a significant difference for the two solvents while the helical angles are in general very consistent with small differences among different modulating angles. While the pitch increases with the modulating angle from 0 to 210.7 µm in IPA and to 149.6 µm in water, the diameter changes less and is saturated at low modulating angles from around 60.0 to 85.1 µm and from 37.0 to 54.0 µm in IPA and water, respectively, with a relative change less than 50%. When the modulating angle is smaller than 15°, the helical structure degrades to a tubular structure. A further decrease of the modulating angle to 0° will result in a ring-like structure with a pitch of 0 µm.

The purpose of having the hard supporting layer is to achieve a stiffness mismatch leading to bilayer deformations. In the meantime, the hard regions serve as directing components that attribute to helical deformations. If the hard regions are strong enough, the direction of the hard regions should be parallel to the axis of the deformed helical tail and therefore the modulating angle and the helical angle are complementary similar to the results in (*16*). However, if they are not strong enough, the hard regions will be pulled by the soft regions and simultaneously form the resulting helical tail. In this case, the hard regions are not only acting as directors but also play a role in the bending motion. Nevertheless, the resulting helical angle still shows a linear decreasing trend in both IPA and water as presented in Fig. 2G.

The helical turns of the deformed helical tails are declining with the increase of modulating angle. For a certain length $L$ of the strip, the turns can be determined by

$$n = \frac{L}{\sqrt{(\pi D)^2 + P^2}}, \tag{4}$$

which means that the helical tail will have fewer turns if it has a larger diameter or pitch. Since both the pitch and diameter are increasing with the modulating angle, a larger modulating angle suggests a smaller number of turns for both in IPA and water. For different modulating angles, there are 40% ~ 50% more number of turns in water than in IPA.

The simulation results of the deformations show a good agreement with the experimental measurements. However, the simulation results of the diameter values for water are a little smaller than the experimental results. This is because of the measurement error in water as the boundary becomes unclear and blurry under a general optical microscope and this could make the measured diameter larger by several micrometers. This also leads to the small deviation of the computed helical angle. For modulating angles smaller than 15°, where the helical tail degrades to a tube or a ring in reality, the self-contact is not considered in the simulations as these degraded structures are not practical microrobot designs. These simulation results will be further used for the prediction of the swimming properties investigated in the next section.

Besides the effects of various modulating angles on the resulting geometry of helical tails, other geometrical parameters such as the effects of the width of the entire strip, the thickness ratio of the soft regions to the overall thickness, and different width ratios of the soft to hard regions were also investigated. Fig. S2 provides experimental and FEA results for tail geometries with different widths. Comparisons with different tail geometries in the Supplementary Information are to the tail design the same parameters as those studied in Fig. 2 with a modulating angle of 45° unless mentioned otherwise. In general, the experimental and simulation results are in good agreement except for the data with a width of 5 μm. The reasons for this deviation can be the weak driving force for deformation and the inaccuracy of fabrication as the feature size (width) becomes smaller. However, both the experimental and simulation results show a significant change in pitch as the width becomes narrower while the diameter and turns are less changed. When the width is larger than 15 μm, the geometrical parameters are very consistent, and the width value is the only difference for the cases in IPA and water.

Unlike the fact that the overall width only affects the pitch, the width of the soft regions will change the pitch, the diameter, and the number of turns (Fig. S3). The pitch varies more when compared to the other two parameters and the trend is opposite to the one with different widths because both increasing the width of the soft regions and decreasing the width of the entire tail will make the relative area of soft regions larger. This will lead to the failure of using hard regions to direct the deformation with a desired helical angle. In general, increasing the ratio of widths of soft and hard regions will make the helical tail more compact with decreased pitch and diameter while the number of turns will be increased. The deviation of soft to hard ratio of 0.75 may also be because of the fabrication inaccuracy where a discrepancy of hundreds of nanometers can lead to a great difference.

The effects of different thickness ratios of the modulating and the supporting layers are studied experimentally with the results provided in Fig. S4. However, the FEA simulations failed to predict the deformations with different thickness ratios with the measured and optimized parameters used for the simulation. The parameters provided good predictions with different in-plane design parameters. However, the out-of-plane direction is more affected by the two-photon polymerization technique where a volume pixel (voxel) with a typical aspect ratio of 3 is observed. Therefore, with a supporting layer of 1 μm, only a couple of layers will be printed, resulting in an overlap of different layers which generates a gradient of the overlapping region where the bottom layer is overlapped for a couple of times and the top layer is only printed once without top overlapping layers. This gradient will promote the deformation and will be reflected in the material parameters used for the simulation. A thicker supporting layer will decrease this gradient phenomenon and the material parameters should be optimized again if needed. However, comparing the experimental results with all design parameters that are investigated, it can be found that the effects of changing the supporting thickness can be achieved by adjusting other parameters and therefore the current material parameters meet the requirements of our studies for the proposed microrobots. Details of the material parameters can be found in the Supplementary Materials.

## Adaptive propulsion of the microrobots

Frequency responsive of the microrobots

Following the tail designs investigated in the last section, different microrobots are fabricated with magnetic heads to obtain magnetic microrobots that can be actuated by a rotating magnetic field. Fig. 3A shows some fabricated microrobots with hydrogel tails with various modulating angles ($\theta$). The geometry of the helical tails attached to the magnetic heads is consistent with the tails we studied in the last section, confirming that the existence of the head will not affect the printing of tails. The microrobots with adaptive tails show different shapes in IPA and water, as expected, while the helical tails are more compact in water. Also, the helical axis of the tails deviates from the axis of the microrobots with a large modulating angle. Only microrobots with a modulating angle greater than 15° are studied with swimming performance since the tails degrade to tubular structures when they are smaller than 15°. A tubular structure will have no net translational propulsion and therefore it is not a practical configuration for swimming. The microrobots are magnetized with a magnetic moment perpendicular to the axis of the heads. Fig. 3B presents the swimming motions of a microrobot swimming at 5 Hz and 120 Hz. As can be seen from the figure, the microrobot has a preferable rotating axis that is not aligned to the helical axis of the tail at a low frequency. However, the rotating axis becomes close to the helical axis at a high frequency as this motion is preferable when the drag force is high (*10*). This result is in agreement with the calculation result of the precession angle shown in Fig. S8B.

For all swimming tests, the microrobots are actuated with a magnetic field of 5 mT. However, the achieved field strength drops to 2.9 mT at 200 Hz due to the limitation of the magnetic actuation system. The velocities of the microrobots are tested near the bottom boundary for convenience since the substrate does not significantly affect the forward velocity (*25*). However, it is inevitable to have drift velocity due to the rotational motion of microrobots on the boundary. The existence of the drift velocity will result in an overall velocity as can be seen in Fig. S6A. The frequency responses of microrobots with different modulating angles in IPA are given in Fig. 3C. As presented in the figure, the larger the modulating angle, the higher the achievable forward velocity. The highest average velocity for a modulating angle of 60° is −4.38 mm/s. The negative value means that the velocity direction is opposite to the direction of the angular velocity of the rotating field. However, the step-out frequency appears earlier as the modulating angle is increased. Therefore, the −4.38 mm/s for a modulating angle of 60° arises at 100 Hz, after which the microrobot exhibits step-out. The frequency responses for those microrobots in water are provided in Fig. S6D. The forward velocities for $\theta = 45°$ and 60° are presented and compared in both IPA and water in Fig. 3D as they have significant forward velocities. For both angles, they show faster forward velocities in IPA than water with a smaller step-out frequency in IPA while the step-out frequency is not observed in water within the frequency range that has been investigated. The maximum velocities and step-out frequencies for different modulating angles are presented in Fig. 3E and Fig. 3F. The maximum velocity decreases with the increase of modulating angle both in IPA and water. The maximum velocities for microrobots in water are obtained at 200 Hz, as the step-out frequency is not found within the test range. In the meantime, the velocities in water for all tested modulating angles are slightly smaller than in IPA. However, testing with a higher frequency will give a faster result as the microrobots are stepped out in IPA at that frequency. The frequency responses of microrobots with different modulating angles in water are given in Fig. S6C for reference. We also found that the drift velocity is diminished as the frequency becomes higher than 100 Hz. Those microrobots show outstanding swimming performance (high forward velocity), as can be seen in Fig. S6D. It is possible that the flow generated at a high frequency will lift the microrobot up so that the interaction between the microrobot and the bottom boundary is weakened making the microrobot act like it is swimming in bulk fluid.

These adaptive microrobots can be applied to different situations based on the desired properties and targeted applications. For example, considering a practical frequency range of 200 and using a microrobot with $\theta = 60°$, the microrobot will only be able to swim in a lower viscosity at 200 Hz. However, using a lower frequency such as 100 Hz, the microrobot will be able to swim in both viscosities. Fabricating the magnetic head with a larger magnetic moment by applying multiple layers of magnetic materials will increase the step-out frequency to above 200 Hz and theoretically will give a much higher maximum velocity since the velocity increases with frequency until step-out. Fig. S6C shows that the magnetic moment increases linearly with the thickness of the magnetic layer. In this case, the microrobot will have a velocity always higher in IPA than in water. Even though the adaption of the fabricated microrobots is tested with IPA and water, which are used for convenience of demonstrating the scheme of applying FEA and dynamic modeling, the tail material is also known for its pH responsiveness which could be used for environmental detection. Moreover, the same fabrication method can be expanded with thermal-responsive hydrogels that make the helical tail deform under the stimulation of temperatures. Assuming using a microrobot with an identical deformation behavior from room temperature to an elevated value to that of from IPA to water, the microrobot will have a compact shape for passing through a narrow channel by sacrificing the velocity. Another example is the adaptation by temperature when applied in biofluids with changes in viscosity. Switching from biofluids with a lower viscosity to a higher one will lead to a step-out and loss of velocity if the microrobot cannot deform to accommodate to the environmental change. Moreover, even if we can lower the frequency to make the microrobot swimmable with a fast forward velocity for some cases, it would be better to keep the frequency to a level higher than 100 Hz since the drift velocity will be minimized. This will benefit applications where boundaries play an important role, such as blood vessels.

Theoretical calculation of the swimming performance

For the different geometrical parameters of the tails and magnetic moment of the head, calculations of the swimming performances of the microrobots are performed. In the theoretical calculation, the two key components are the magnetic moment and the mobility tensors that correspond to a certain geometry. The magnetic moment is measured during the magnetization process while the mobility tensors are obtained via calculations by discretizing the microrobot into spheres. Since the magnetic heads are made with permanent magnetic materials and magnetized perpendicular to the axis of the heads, the magnetic moment is consistent and affixed to the body of the microrobot. Details of the calculations can be found in the Supplementary Materials. However, the method is scaled with a factor of 10 to adjust the characterization frequency by using a magnetic moment of $2.8\times10^{-7}$ emu instead of the measured value of $2.8\times10^{-6}$ emu (Fig. S7). The scaled method is also used previously and shows good agreement with the experimental results (*16*). A new perspective on this is that the theory of calculating the dynamics of microrobots does not hold at the high frequency where the swimmers are not at low Reynolds numbers. An estimation based on the size of the microrobots gives a Reynolds number of 0.74 and 2.26 for IPA and water, respectively, with a diameter of 60 μm at 200 Hz. Obviously, even though these values are small, they are not small enough to ignore the inertial terms and this will make the prediction inaccurate in the high frequency regime. The rotational motion of the microrobot at a high frequency will generate a strong flow around it. However, a factor of 10 can compensate for this discrepancy at least for the frequency regime below 200 Hz. This is proved in Fig. 3F as the measured step-out frequencies, considering both the frequency response with the head of the microrobot sharing an identical (Fig. 3C) and opposite (Fig. S6C) direction to the velocity direction, are in good agreement with the predicted results where an increment in modulating angle will result in a decrease in the step-out frequency. Definitions of the existence of these opposite velocities can be found in Fig. S6A. The main reasons causing the deviations between the experimental and calculation results, reflected mainly on the slopes for different microrobots, are the miss-alignment of the tail and head axes and the out-of-plane bending. These are not considered in the calculation since the effects are small and will not significantly affect the overall results.

Fig. 3E and Fig. 3F conclude the prediction results for microrobots fabricated with different modulating angles. Even though the step-out frequency decreases with the increase of the modulating angle, the maximum forward velocity is increased because the helical tail is more compact with a smaller modulating angle which makes it less preferable for swimming giving a less significant slope. For all the studied geometrical parameters (as shown in Fig. 3E, Fig. 3F, Fig. S9, and Fig. S10), the forward velocity is always faster in IPA at the step-out frequency than in water at 200 Hz. However, the velocity for water at the step-out frequency is higher than in IPA. Currently, the maximum velocity for IPA is fixed at a step-out frequency that is less than 200 Hz while the velocity for water still has the potential to increase.

Fig. S9 discusses the effects of the width of the tail on the swimming performance while Fig. S10 analyzes the effects from different ratios of soft to hard regions. Having a wider width will decrease both the step-out frequency and the velocity at that frequency. However, while the step-out frequency has a smooth change for all widths investigated, the maximum velocity shows a significant drop when the width is 5 μm. For different widths, it concludes that the increase in width will make the swimming performance worse with a smaller step-out frequency and a slower velocity. This result is in agreement with the result from Morozov and Leshansky (*26*). However, since a soft tail with a smaller width will be more fragile, a balance is needed since a smaller width will give a higher velocity. For different soft-to-hard ratios, increasing the ratio will increase the step-out frequency but drop the maximum velocity.

An inverse strategy for achieving target microrobots

An inverse strategy for achieving microrobots for specific applications is proposed, as shown in Fig. 3G. The material properties of the hydrogels are experimentally measured for use with FEA studies of the helical tail deformations. Using the geometrical parameters of the deformed helix obtained from the FEA results, theoretical calculations are performed to investigate the swimming performance of the fabricated microrobots. In the previous two sections, both the FEA simulations and the theoretical calculations were verified with experiments that prove the feasibility of using the combination of the two theoretical methods in different environments. Therefore, one can use the proposed strategy to predict the swimming performance in various environments to fabricate the desired microrobots as long as the material properties are known. Since the chain number $Nv$ and stretch at the dry state $\lambda_0$ are not consistent in different environments, the interaction parameter $\chi$ can be estimated based on the size of the hydrogel in the corresponding environment. The $\chi$ values used for the soft regions are obtained based on the designed size and the size in IPA while the material properties are measured in water.

**Passive and Active Adaptive Gap Traversing**

To realize application tasks such as adaptive locomotion and micromanipulation, the controlled propulsion of the microrobots is first tested. The testing platform is shown in Fig. 4A. The system mainly consists of a magnetic field generator, a microscope, and a camera. An extra light source is used to provide appropriate imaging quality. Fig. 4B shows controlled propulsion with "P" and "U" trajectories, representing Purdue University. The trajectories were obtained manually by tuning the direction of the angular velocity of the rotating magnetic field with rotating frequencies of 160 Hz for "P" and 120 Hz for "U". As shown in the images, the microrobots are able to swim with their axes aligned with the paths due to the minimization of the drift velocity. However, imperfections and/or debris on the substrate will introduce disturbances. Nevertheless, the trajectories can be reestablished and, in general, give recognizable patterns.

Two types of adaptive locomotion are demonstrated next. The application here is for traversing confined gaps within a microchannel. A microchannel with a wider entrance but a narrower outlet is designed to perform the real-time deformation of the adaptive microrobot. Due to the soft nature of the helical tail and the deformation period, the microrobot can pass through the confined gap during the deformation process following the solvent flow. Fig. 4C demonstrates this passive adaptive locomotion behavior of the proposed microrobot. The process is illustrated at different time frames. The microrobot is initially in an IPA environment. At $t = 5.0$ s, water is introduced and the microrobot has flowed towards the gap with its original IPA-shape. Since the microrobot is wider than the gap, a rigid microrobot would not be able to pass through it. However, the soft nature of the hydrogel tail allows our microrobot to locomote through the narrow gap. As the water reaches the gap at $t = 5.5$ s, the microrobot traverses the gap at the same time as it is deforming due to the presence of a water environment. During the period from $t = 6.7$ s to $t = 17.4$ s, the microrobot is completely out of the gap and has completed its deformation.

Rather than taking advantage of its soft nature and the deformation process, the gap traversing can be achieved by a more active approach. In a situation illustrated as in Fig. 4D, the microrobot is in its loose form initially and is not able to pass the gap. However, it will deform into a compact form when it is in water. By completing the shape-change process before entering the confined gap, the microrobot can be actively steered to pass through the gap. Fig. 4E demonstrates this process. Baffle features have been added to the test fixture to prevent the microrobot from passively traversing the gap due to the water flow. As can be seen in the figure, the microrobot is first controlled to move around the first baffle from $t = 0$ s to $t = 2.9$ s at 5 Hz. Note: the microrobot is swimming tail-forward.

Then, sufficient water is introduced into the environment to trigger the deformation into the compact form. At $t = 22.9$ s, the microrobot completes its deformation and is ready to continue swimming. During the period from $t = 22.9$ s to $t = 37.5$ s, the microrobot is successfully maneuvered into the gap at 11 Hz. At $t = 42.5$ s, the microrobot is completely through the gap.

**Targeted Micromanipulations and Potential for Biomedical Applications**

To fully explore the potential of this multi-material microrobot, another material and/or structure can be introduced on the other side of the head, opposite to the tail, to provide additional functionalities. By printing a holder onto the head of the microrobot using a double fish-bone design, the microrobot now has two advanced functionalities, e.g., adaptive locomotion and micromanipulation capabilities. The holder is able to enclose an object when the microrobot is pushing along the head direction or catch an object with a fingered design. Fig. 5A illustrates a micromanipulation task using the microrobot to transfer a spherical object from one side of the microchannel to the other side. By applying an inverse direction of the rotating field, the object can be released. A demonstration is realized in Fig. 5B successfully manipulating a cluster of spheres from the right side of the microchannel to the left side. The actuation starts at position (i) and approaches the targeted object at (ii). After some manual controls, the cluster is picked up. Then, the microrobot is controlled from position (iii) to (v) to enter the channel connecting the right and left sides. After passing the channel to position (vii), the manipulation is stopped. Finally, a reverse field is applied to release the object. The released cluster is stationary and the final position of the microrobot is shown in (viii).

With the holder design, the microrobot is able to perform targeted delivery. Assuming the delivered object is loaded with drugs, targeted drug delivery can be achieved with such a microrobot. Even though the microrobot is tested with IPA and water, the hydrogel tail is also responsive to heat or different pH values (*27*). To verify that our microrobots are biocompatible for potential biomedical applications, cells are grown on the individual materials that were used to fabricate the microrobots and with some fabricated microrobots to see their viability with those materials. Fig. 5C shows the cells grown with the SU-8/MMPs composite after incubating for one day. In this image, only cells that are grown next to the material are featured since the SU-8/MMPs composite is too dark to visualize the cells grown on it. The living cells demonstrate the noncytotoxic properties of the material. Fig. S13 shows cells grown on all other materials used to fabricate the microrobot with a bare glass substrate as a control group. The growing cells prove that the materials are non-cytotoxic and are in agreement with the results in other studies (*28–31*). Additionally, the cells can proliferate directly onto a fabricated microrobot, as seen in Fig. 5D, further demonstrating the microrobot's biocompatibility. Moreover, scanning electron microscope (SEM) images are obtained with multiple microrobots showing cells directly grown on the helical tail, magnetic main body, and functional end-effector, respectively, as shown in Fig. 5E and Fig. S14. Even though the microrobot is not anticipated for continuous application of more than 1 day, the cell viability of cells incubated with the materials for 3 days are also studied. As can be seen in Fig. S15, the results show that the materials used to make the microrobot have no major cytotoxicity on cells even after 3 days. These results prove that these microrobots are viable for future biomedical applications such as cell manipulation, tissue engineering, and drug delivery.

# CONCLUSION

This paper demonstrates a complete investigation of a shape-programmable adaptive helical microrobot, that consists of multiple materials, from the perspective of both simulation and experiments. Material properties were measured so that FEA simulations could be performed to study the adaptation of helical hydrogel tails in different environments. With different geometrical designs of the helical tails, the shapes in various environments can be predicted and therefore be programmed. Many geometrical parameters were simulated and verified by experiments with fabricated hydrogel tails. Meanwhile, the variations of swimming performance upon environmental changes of the microrobots were tested to achieve a programmable swimming performance. The swimming performance for a microrobot with a prescribed initial geometry swimming in different environments was successfully predicted using the combination of these two numerical methods and validated by experiments. Demonstrations of applying adaptive locomotion were successfully performed both passively and actively. Moreover, a second advanced function was introduced into the microrobot through the addition of an end-effector for micromanipulation tasks. Tests were conducted demonstrating the potential of using such adaptive microrobots for targeted delivery. Overall, the investigations performed in this paper provide a standardized systematic strategy for studying adaptive microrobots while also introducing a multi-material microrobot with two advanced functionalities for potential biomedical applications.

# MATERIALS AND METHODS

## Materials

Acrylic acid (AAc), polyvinylpyrrolidone (PVP), ethyl lactate (EL), triethanolamine (TEA), *4,4'*-bis(diethylamino)benzophenone (EMK), and *N,N*-dimethylformamide (DMF) are purchased from Sigma-Aldrich. *N*-Isopropylacrylamide (NIPAM) and dipentaerythritol pentaacrylate (DPEPA) are obtained from Scientific Polymer Products Inc. and Alfa Chemistry, respectively. The photoresist SU-8 2025 is purchased from Fischer Scientific. The neodymium magnetic microparticles (MMPs, MQFP-B) with an average diameter of 5 μm are obtained from Magnequench. The TPP resin IP-S and IP-Q are obtained from Nanoscribe GmbH. The adhesion promoter 3-(Trimethoxysilyl)propyl methacrylate (TMPSM) is obtained from Acros Organics. Polydimethylsiloxane (PDMS) is purchased from Electron Microscopy Sciences. Cells from African green monkey kidney (COS-7), Dulbecco's Modified Eagle's Medium (DMEM), and fetal bovine serum (FBS) are obtained from American Type Culture Collection. Penicillin-streptomycin, trypsin-EDTA, and phosphate-buffered saline (PBS) are purchased from Gibco. The trypan blue dye is obtained from VMR. All materials are used as obtained without further modifications.

## Experimental platform

The printed microrobots are actuated by a rotating magnetic field generated by the Magnebotix field-generating system (MFG-100i, Magnebotix, Switzerland) with a microscope placed above the system. The magnetic field generator is able to provide rotating magnetic fields in three-dimensional space. The system is able to generate magnetic fields with a field strength of up to 20 mT and a rotating frequency of up to 500 Hz. However, 5 mT is typically used with a maximum frequency of 200 Hz in this paper, considering the heating effects of the system for long-term tests.

## Preparation of the hydrogel precursor and SU-8/MMPs mixture

The hydrogel precursor and the SU-8/MMPs mixture are prepared following the procedure described in (*16*). Briefly, a monomer solution is obtained by dissolving 1.6 g of NIPAM, 0.8 mL of AAc, and 0.15 g of PVP into 1 mL of EL. After complete dissolution, 2.5 mL of the above solution is added with 0.4 mL of DPEPA, 0.5 mL of TEA, and 100 μL of EMK/DMF solution with a weight ratio of 0.2. The SU-8/MMPs mixture is prepared by mixing the two riwith a weight ratio of 1 with a high-speed mixer at 7000 rpm.

**Fabrication of the adaptive helical tails for parametric studies**
The parametric studies of the adaptive tails are investigated without magnetic materials; therefore, the helical tails are printed with IP-S fixtures to keep the tails from floating away. The IP-S fixtures are first printed on a glass coverslip with a diameter of 30 mm using a commercial TPP system (Photonic Professional GT2 (PPGT2), Nanoscribe GmbH). The hydrogel tails are then printed with a 63× objective with the dip-in configuration. To realize a printing process with the dip-in configuration, a droplet of immersion oil is applied on top of the coverslip. Since the hydrogel precursor is on the bottom of the coverslip, this technique effectively increases the thickness of the coverslip in order to realize the interface-finding function of the TPP system. Details of this setup are provided in Fig. S11.

**Fabrication of the adaptive multi-material microrobots**
The schematic of the fabrication process can be found in Fig. S5. Briefly, a SU-8 bottom layer is first patterned followed by a layer of SU-8/MNPs mixture on a 30 mm coverslip. Then the fish-bone layer is applied on top of the magnetic layer on a coverslip substrate. Before developing the fish-bone layer, an identical design as the bottom is patterned along with the fish-bone design to enclose the magnetic part with pure SU-8 to ensure biocompatibility. For the microrobots with a holder end-effector design, a double fish-bone design is adopted. The holder is 3D-printed with IP-S with a 25× objective using the PPGT2. The obtained structures are then magnetized with a preferred direction of the magnetic moment with a 9 T magnetic field using a Physical Properties Measurement System (PPMS, Dynacool, Quantum Design). Then the substrate is transferred back to the TPP system for the printing of the hydrogel connector and tail using a 63× objective. After the printing, the structures are developed in IPA for an hour before being immersed in water. For all printing processes, immersion oil is applied on the other side of the substrate to increase the effective thickness.

**Mechanical properties testing**
The mechanical properties of the hydrogels with different crosslinking densities are obtained by performing uniaxial compression tests on prism-shaped test samples with a dimension of 40 μm×40 μm×100 μm with a custom-built testing platform. To increase the adhesion of the samples for testing, a TMPSM adhesion layer is self-assembled onto an indium tin oxide-coated substrate by submerging the substrate into 100 mL of acetone with 250 μL of TMPSM. Details of the experimental setup can be found in the Supplementary Materials.

**Numerical analysis**
The FEA is performed by the commercial FEA software ABAQUS with a user-defined subroutine UHYPER, which is used to define isotropic hyperelastic materials with their energy function and the corresponding derivatives over different invariants. Modeling details can be found in the Supplementary Materials. The final form of the free energy function to be implemented in UHYPER subroutine is given as:

$$W = \frac{1}{2} N v \left[ J^{\frac{2}{3}} \bar{I}_1 - 3 - 2\log(J) \right] + (J-1) \left[ \log\left(\frac{J-1}{J}\right) + \frac{\chi}{J} \right] - \frac{\mu}{kT}(J-1). \quad (5)$$

**Dynamic modeling**

The translational velocity can be obtained by applying the dot product of the translational and the angular velocities (*12, 32*), which is

$$U_Z \cdot \omega \cdot l = \mathbf{\Omega}^{BCS} \cdot \boldsymbol{\mathcal{G}} \cdot \boldsymbol{\mathcal{F}}^{-1} \cdot \mathbf{\Omega}^{BCS}, \tag{6}$$

where $l$ is the characteristic dimension of the microrobot, $\boldsymbol{\mathcal{G}}$ and $\boldsymbol{\mathcal{F}}$ are the coupling and rotational mobility tensors of the microrobot achieved by the multipole expansion method by discretizing the obtained microrobots into spheres (*33*). More detailed information can be found in the Supplementary Materials.

**Fabrication of the microchannels**

For different tests of the microrobots, multiple microchannels are prepared using PDMS. While the chambers for the tests of swimming performance are obtained by puncturing a circular hole into a PDMS film, the microchannels with more complex designs for adaptive locomotion and micromanipulation tasks are achieved via a soft lithography process. The channel designs are printed via TPP with a 10× objective using the IP-Q resin on a silicon substrate. Uncured PDMS is applied on top of the IP-Q structure to get the channels. Larger test spaces can be achieved by puncturing holes with the obtained microchannels.

The environment used to perform the adaptive locomotion demonstrations is prepared differently. The structure of the confined gap is printed with IP-S on a 30 mm coverslip with the dip-in configuration with immersion oil on the other side to increase the effective thickness for printing. Then the coverslip with a printed test feature is used to prepare the microchannel for adaptation tests by coverslips as spacers which are enclosed by another coverslip, as shown in Fig. S12.

**Cytotoxicity Testing and SEM Imaging**

The samples with individual materials are prepared by spin-coating a layer of such material on a glass substrate (coverslip of 30 mm diameter) while the samples with microrobots are used after a normal fabrication process as demonstrated above. All materials are immersed in IPA overnight and baked to remove any residue solvents after preparation. Three samples of each material are prepared and sterilized under UV light for 1 hour.

The COS-7 cells are maintained in a growth medium composed of DMEM with 10% FBS and 1% penicillin-streptomycin and cultured at 37°C with 5% $CO_2$. Then the COS-7 cells ($4\times10^4$ cells/ml) are seeded onto the materials in a 24-well plate with growth medium and incubated for 24 hours at 37°C with 5% $CO_2$ again. Brightfield images are captured using an inverted light microscope after 1 day of incubation.

Another batch of samples is prepared with the same process described above for 3 days in growth medium at 37 °C with 5% $CO_2$. After 3 days, cells were harvested using trypsin-EDTA and resuspended in growth medium to create cell suspensions. The cell suspension was mixed with 0.4% trypan blue dye at a 1:1 (v/v) ratio. For each sample, 20 μL of stained cells was pipetted into a hemacytometer (Bright-Line), and viable/dead cells were counted under an inverted light microscope. Each sample was drawn and counted twice.

The samples used for SEM imaging are prepared following the process provided in (*10*). In brief, the samples are incubated for one day and then are gently rinsed by PBS twice followed by a two-step fixation using 3.7% formaldehyde in PBS and 2.5 % glutaraldehyde in PBS for 10 minutes and overnight, respectively. The samples are then rinsed in fresh PBS and kept at 4°C before drying. The samples are dried by a critical point dryer (Autosamdri-931, tousimis). To do so, the samples in PBS are processed by a gradient of ethanol of 10%, 25%, 50%, 75%, 90%, and 100% in water. After drying, the samples are sputtered with a 40 nm layer of gold/palladium (60:40). The SEM images are taken by a Hitachi S-4800 SEM.

**Supplementary Materials**
    Material property testing
    Finite element analysis
    Dynamic calculation of the microrobots
    Fig. S1 to Figs. S15
    References (*34–37*)
    Movie S1 to Movies S3

**Acknowledgments:** The authors would like to thank Prof. Adrian Buganza Tepole in School of Mechanical Engineering at Purdue University for providing access to ABAQUS. The authors would also like to thank Siting Zhang and Prof. Luis Solorio in Weldon School of Biomedical Engineering at Purdue University for discussions on cytotoxicity testing. The authors would also like to thank Prof. Fang Huang in Weldon School of Biomedical Engineering at Purdue University for providing cell lines to initiate the cytotoxicity testing.

**Funding:** This work was supported by the National Science Foundation (NSF IIS Award 1763689 and NSF CMMI Award 2018570).

**Author contributions:** L.T. designed the overall technical path of the paper. L.T. built the test platform and performed the material tests. L.T. fabricated tails and completed geometrical studies. L.T. completed the finite element analysis. L.T. and Y.Y. fabricated the microrobots and performed the velocity tests. L.T. performed the theoretical calculation of swimming performance prediction. L.T. performed the application tasks. L.F. worked on the cytotoxicity tests. D.J.C obtained funding for and advised the entire project. L.T. drafted the manuscript. D.J.C and L.T. reviewed and revised the manuscript. All authors have access to all the data and revised and approved the final manuscript.

**Competing interests:** Authors declare that they have no competing interests.

**Data and materials availability:** All data needed to support the conclusions of the paper are available in the paper or the supplementary materials.

**Figures:**

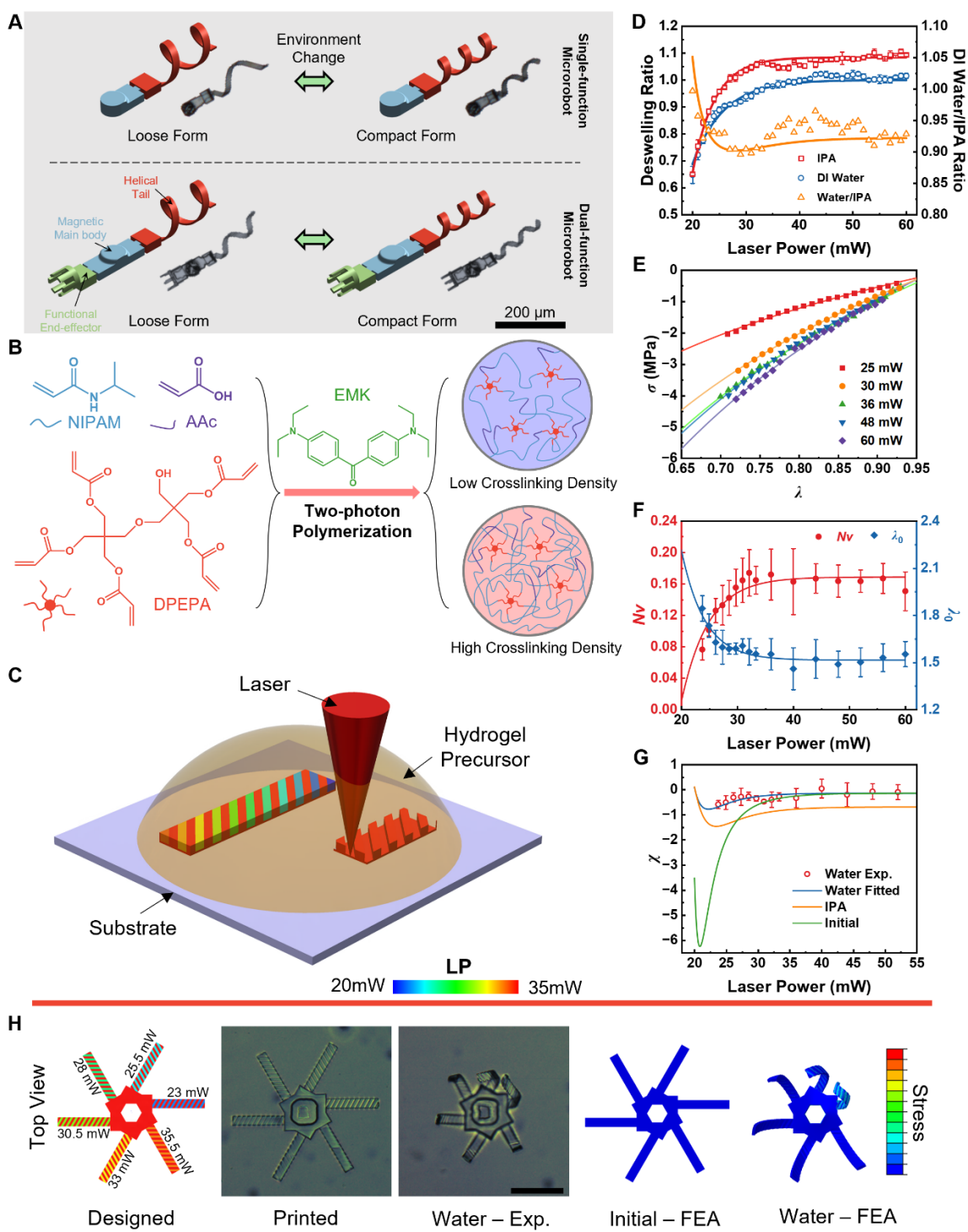

**Fig. 1. Schematic of the object delivering microrobots and material properties of the hydrogels for FEA simulation.** (**A**) Schematic of the shape-adaptive microrobots for targeted delivery. (**B**) Schematic of the two-photon polymerization process with laser power-controlled crosslinking density. (**C**) Schematic of the printing process with different laser powers. (**D**) Deswelling ratios of the hydrogels in DI water and IPA with different laser powers and the corresponding ratio of deswelling ratios in the two solvents. (**E**) Stress-strain curves of samples printed with different laser powers. (**F**) Crosslinking densities ($Nv$) and the relative stretch ($\lambda_0$) of samples achieved by different laser powers. (**G**) Effect of laser power on the Flory interaction parameters ($\chi$). Based on the deswelling ratios with respect to the designed size, the

$\chi$ values can be calculated from the corresponding $\lambda_0$ while the $Nv$ is consistent. (**H**) Experimental and simulation results of a flower-like structure achieved by soft regions printed with different laser powers. Scale bar: 100 μm.

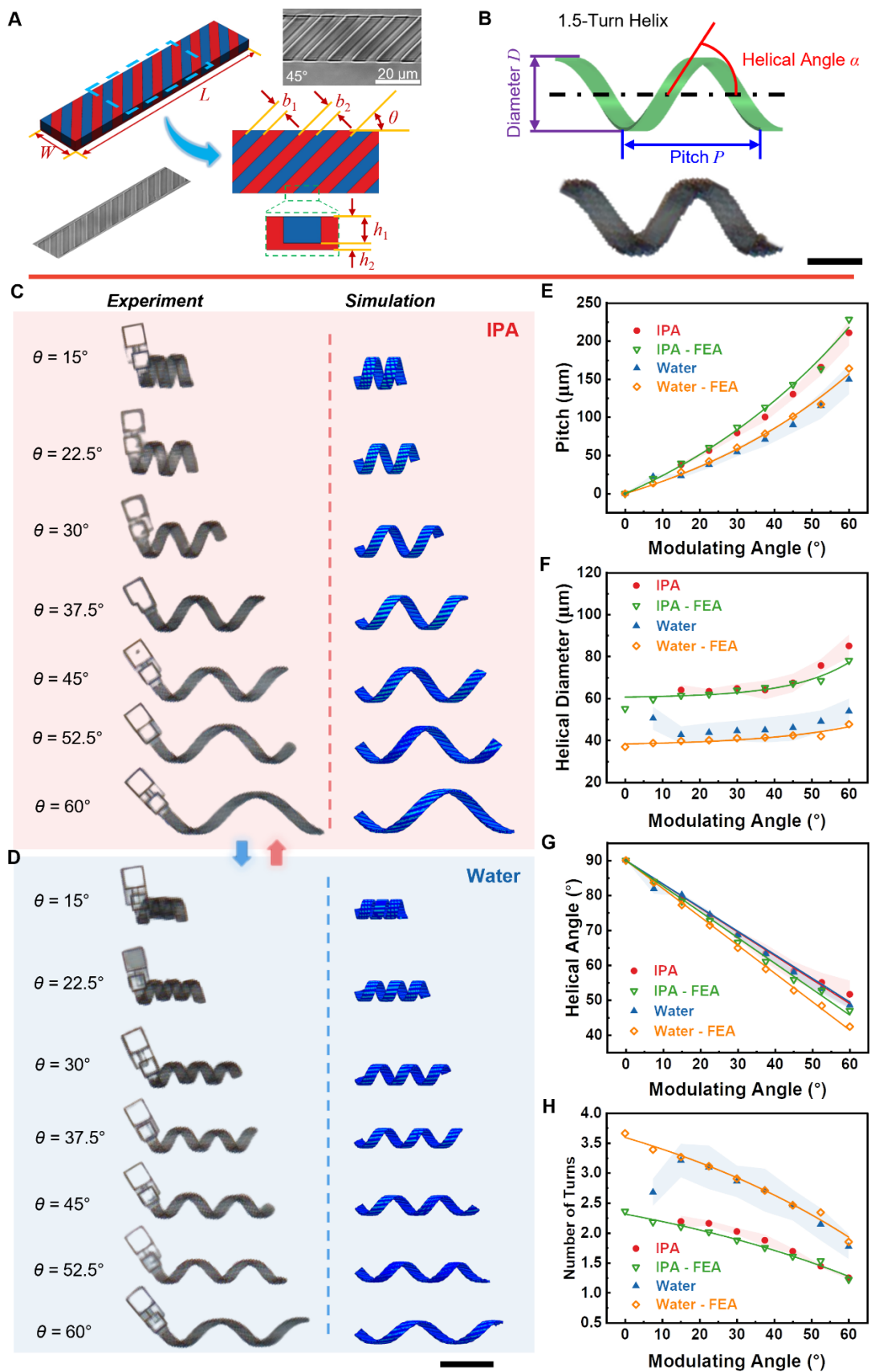

**Fig. 2. Geometrical design of the helical tails.** (**A**) Design parameters during the fabrication process. (**B**) Illustration of the geometrical parameters of a helical structure. (**C**) Helical deformation of the tails with different modulating angles ($\theta$)

in IPA. (**D**) Helical deformation of the tails with different modulating angles ($\theta$) in water. Scale bars in (**C**) and (**D**): 100 μm. (**E**) Measured and simulated pitch ($P$) with different modulating angles ($\theta$) in water and IPA. (**F**) Measured and simulated helical diameter ($D$) with different modulating angles ($\theta$) in water and IPA. (**G**) Measured and simulated helical angle ($\alpha$) with different modulating angles ($\theta$) in water and IPA. (**H**) Calculated and simulated pitch ($P$) with different modulating angles ($\theta$) in water and IPA. Shaded regions in (**E**) to (**H**) are error bars of standard deviations obtained from five samples.

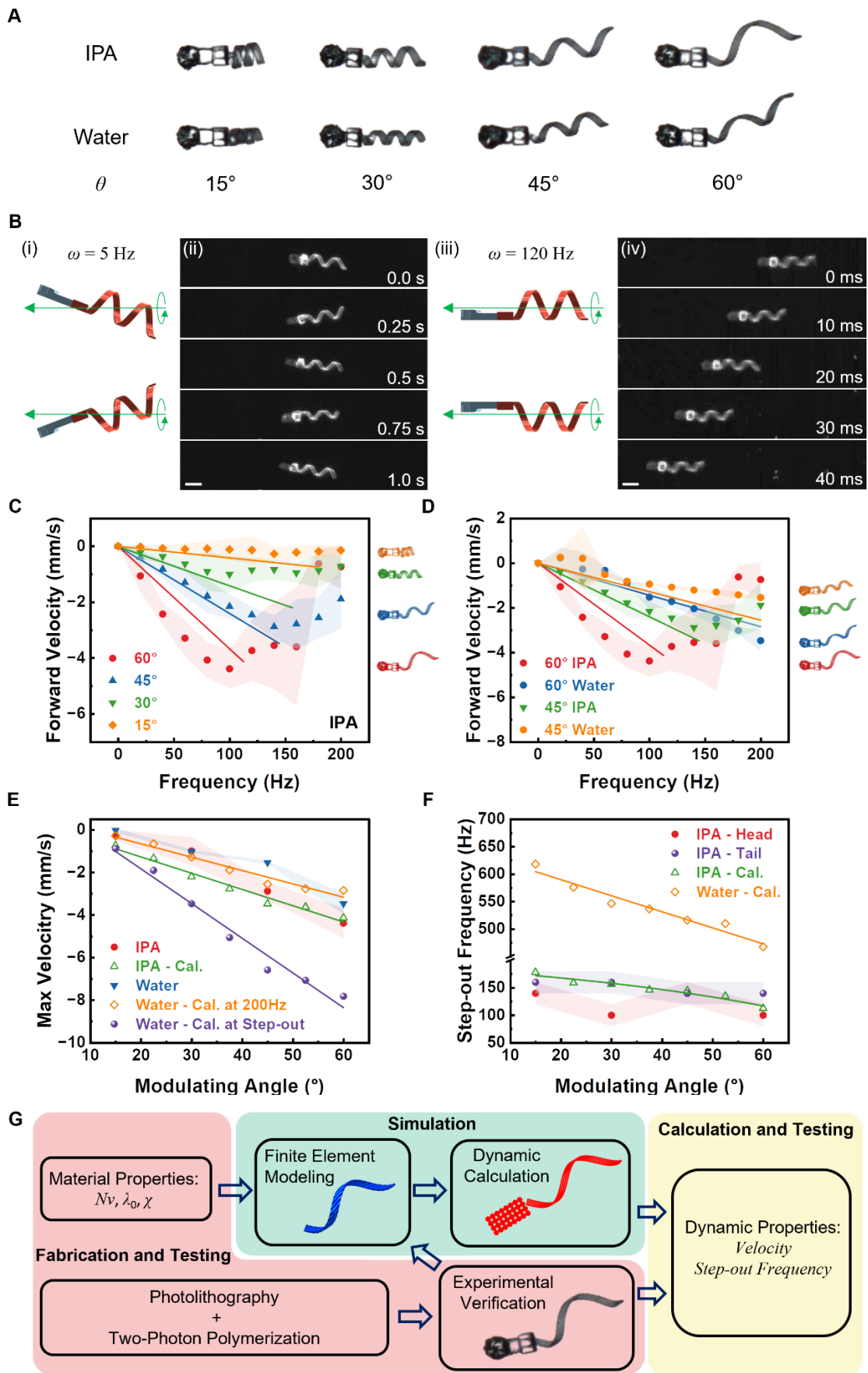

**Fig. 3. Adaptive microrobots and adaptive propulsion.** (**A**) Fabricated microrobots with different modulating angles in IPA and water. (**B**) Swimming motion of the microrobot at low (5 Hz) and high (120 Hz) frequencies. Scale bar: 100 μm. (**C**) Forward velocities under different frequencies of microrobot with various modulating angles in IPA. (**D**) Comparison of forward velocities in IPA and water of microrobots with modulating angles of 45° and 60°. (**E**) Highest forward velocities and (**F**) step-out frequencies of microrobots with different modulating angles in IPA and water. The head and tail in the legend represent the two swimming types as illustrated in Fig. S6A. Shaped regions in (**C**) to (**F**) are error bars of standard deviations obtained from five samples. (**G**) The inverse strategy for achieving targeted microrobot behavior by the combination of FEA and dynamic calculations.

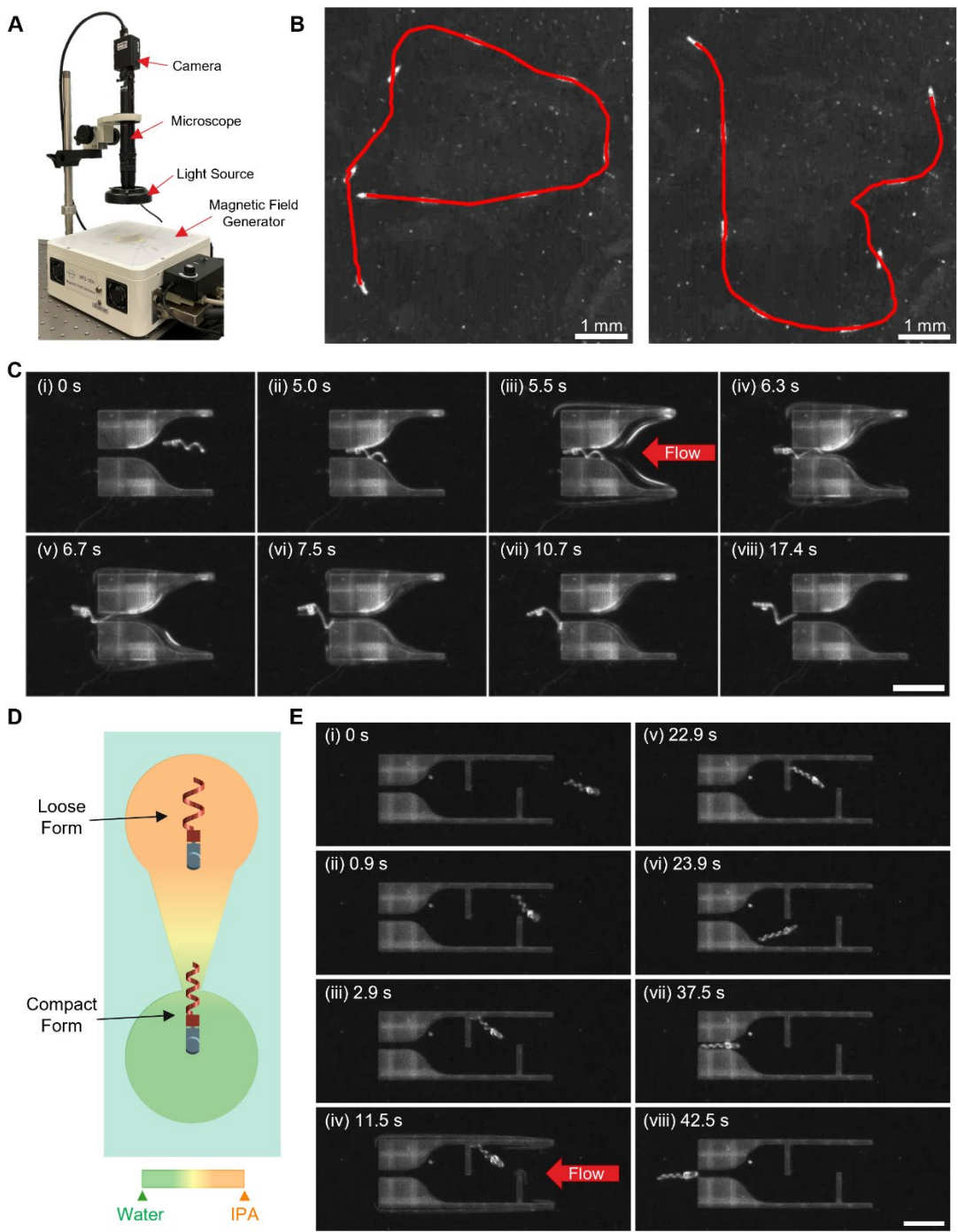

**Fig. 4. Controlled propulsion and adaptive locomotion.** (**A**) Control platform for microrobot actuation with visual feedback. (**B**) Controlled propulsion achieving "P" and "U" trajectories using a microrobot. (**C**) Passive adaptive behavior for traversing a confined gap. Scale bar: 400 μm. (**D**) Illustration of actively adaptive behavior for passing through a confined gap using a microrobot. (**E**) Experiment demonstration of active adaptive behavior for traversing a confined gap. Scale bar: 400 μm.

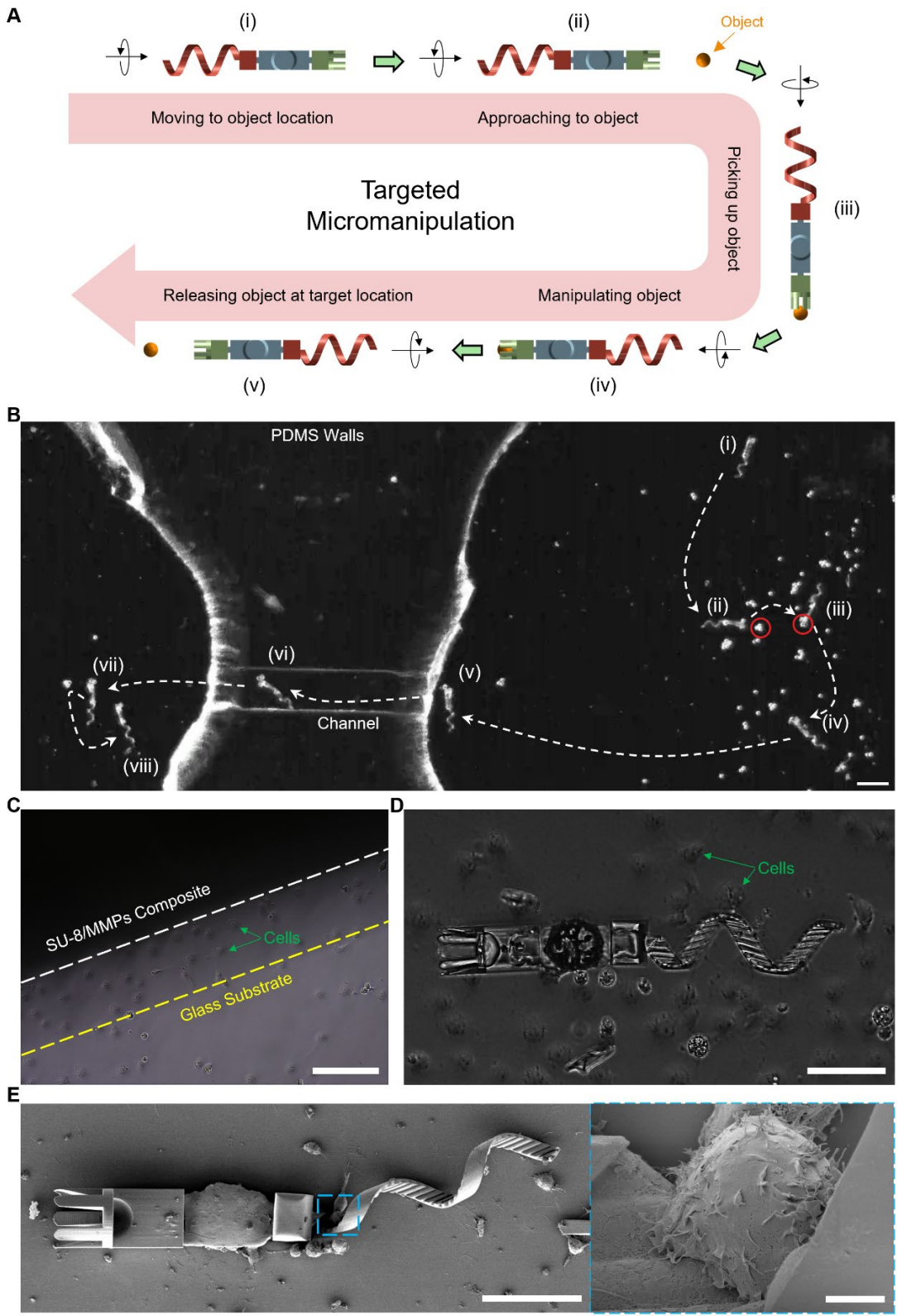

**Fig. 5. Micromanipulation for targeted object delivery and biocompatibility.** (**A**) Schematic of using the microrobot for micromanipulation. The process can be separated into five main steps: (i) moving the microrobot to the object location, (ii) approaching the target object, (iii) picking up the object, (iv) transporting the object, and (v) releasing the object at targeted delivering location. (**B**) Micromanipulation realized by the microrobot with a functional holder moving a cluster of spheres from one side of the channel to the other. Scale bar: 200 µm. (**C**) Cells grown with SU-8/MMPs composite which is used to fabricate the magnetic head of the microrobots after 1-day incubation. Scale bar: 200 µm. (**D**) Cells grown with microrobot with a functional end-effector. Scale bar: 100 µm. (**E**) SEM image of a microrobot that has cells grown on it. Scale bar (left): 100 µm. The image on the right is an enlarged image of the blue highlighted region in the left image, showing a cell grown directly on the helical tail of the microrobot. Scale bar (right): 5 µm.